%% file: acl_latex.tex
\pdfoutput=1
\documentclass[11pt]{article}
\usepackage[]{acl}
\usepackage{algorithm}
\usepackage{algorithmic}
\usepackage{listings}
\usepackage{times}
\usepackage{graphicx}
\usepackage{fancyvrb}
\usepackage{latexsym}
\usepackage{multirow}
\usepackage{xspace}
\usepackage{booktabs}
\usepackage{graphicx}
\usepackage{bbding}
\usepackage{multirow}
\usepackage{booktabs}
\usepackage{balance}
\usepackage[T1]{fontenc}
\usepackage[utf8]{inputenc}
\usepackage{amssymb}
\usepackage{amsmath}
\usepackage{bm}
\usepackage{microtype}
\usepackage{array}
\usepackage{listings}
\newcommand{\tkgqa}{TKGQA\xspace}

\newcommand{\cronq}{\textsc{CronQuestions}\xspace}

\newcommand{\ourq}{\textsc{MultiTQ}\xspace}

\newcommand{\embed}{\textsf{EmbedKGQA}\xspace}

\newcommand{\tempqr}{\textsf{TempoQR}\xspace}
\newcommand{\eae}{\textsf{EaE}\xspace}
\newcommand{\cronkgqa}{\textsf{CronKGQA}\xspace}
\newcommand{\multitq}{\textsf{MultiQA}\xspace}

\newcommand{\chatgpt}{\textsf{ChatGPT}\xspace}
\newcommand{\rag}{\textsf{KG-RAG}\xspace}
\newcommand{\chainofthought}{\textsf{CoT}\xspace}
\newcommand{\react}{\textsf{ReAct}\xspace}

\newcommand{\ourmodel}{\textsf{ARI}\xspace}

\newcommand{\myparagraph}[1]{\vspace{1ex}\noindent\textbf{#1.}\hspace{1em}}
\title{Temporal Knowledge Question Answering \\ via Abstract Reasoning Induction}

\author{
    \begin{tabular}{c}
    Ziyang Chen$^{1}$\footnotemark[1]~ \quad 
    Dongfang Li$^{2}$\footnotemark[1]~ \quad 
    Xiang Zhao$^{1}$\footnotemark[2]~ \quad
    Baotian Hu$^2$\footnotemark[2]~ \quad 
    Min Zhang$^2$ \vspace{.5mm} \\
    \end{tabular}
    \\ \vspace{.5mm}
    \begin{tabular}{c}
    $^1$ Laboratory for Big Data and Decision, National University of Defense Technology, China\\
    $^2$ Harbin Institute of Technology (Shenzhen), Shenzhen, China \\
    \end{tabular}
    \\ \vspace{.5mm}
     \texttt{\{chenziyangnudt,xiangzhao\}@nudt.edu.cn} \\
    \texttt{\{lidongfang,hubaotian,zhangmin2021\}@hit.edu.cn} \\  
}

\begin{document}
\maketitle
\renewcommand{\thefootnote}{\fnsymbol{footnote}}
\footnotetext[1]{Equal Contribution.}
\footnotetext[2]{Corresponding authors.}

\begin{abstract}
\input{chapters/abstract}

\end{abstract}

\section{Introduction}
\input{chapters/introduction}

\section{Related Work}
\input{chapters/related}


\section{Method}
\input{chapters/method}

\input{chapters/experiment}

\section{Conclusion and Limitation}
\input{chapters/conclusion}

\section*{Acknowledgement}
\input{chapters/acknowledgement}

\bibliography{custom}

\appendix

\section{Appendix}
\input{chapters/appendix}


\end{document}

%% file: chapters/abstract.tex
In this study, we address the challenge of enhancing temporal knowledge reasoning in Large Language Models (LLMs). LLMs often struggle with this task, leading to the generation of inaccurate or misleading responses. This issue mainly arises from their limited ability to handle evolving factual knowledge and complex temporal logic.
To overcome these limitations, we propose Abstract Reasoning Induction (\ourmodel) framework, which divides temporal reasoning into two distinct phases: Knowledge-agnostic and Knowledge-based. This framework offers factual knowledge support to LLMs while minimizing the incorporation of extraneous noisy data. Concurrently, informed by the principles of constructivism, 
\ourmodel provides LLMs the capability to engage in proactive, self-directed learning from both correct and incorrect historical
reasoning samples. By teaching LLMs to actively construct knowledge
and methods, it can significantly boosting their temporal reasoning abilities.
Our approach achieves remarkable improvements, with relative gains of 29.7\% and 9.27\% on two temporal QA datasets, underscoring its efficacy in advancing temporal reasoning in LLMs. The code can be found at \url{https://github.com/czy1999/ARI-QA}.


%% file: chapters/introduction.tex
\vspace{10pt}
\begin{center}
\begin{tabular}{p{0.78\linewidth}}
\textit{"Knowledge is not simply transmitted from teacher to student, but actively constructed in the mind of the learner."}
\vspace{5pt}

--- Jean Piaget
\raggedleft





\end{tabular}
\end{center}
\vspace{-5pt}

\begin{figure}[t]
  \centering
  \includegraphics[width=\linewidth]{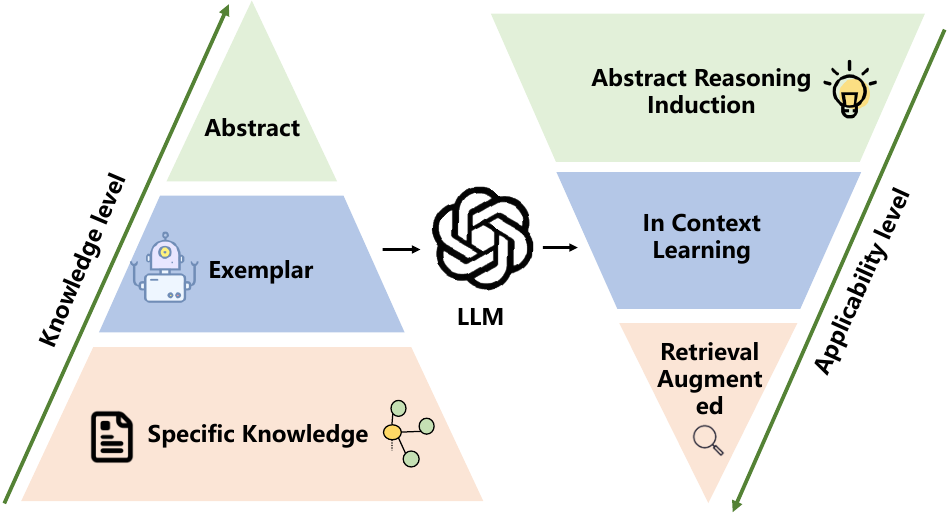}
  \caption{LLMs, when integrated with various levels of information, exhibit varying scopes of applicability; the more abstract and refined the knowledge, the broader its potential application.}
  \label{fig:intro}
  \vspace{-10pt}
\end{figure}

In practical scenarios, factual knowledge frequently undergoes evolution over time~\cite{DBLP:journals/tkde/RoddickS02, DBLP:conf/www/HoffartSBLMW11, liang2023learn,Liangke_Survey}. For instance, the host city of the Winter Olympic Games in 2018 was South Korea, while in 2022 it was Beijing. Despite their proficiency in a range of linguistic tasks, LLMs often demonstrate limitations in the efficient processing and understanding of tasks that involve temporal information~\cite{DBLP:conf/acl/0009C23, DBLP:journals/corr/abs-2303-18223,Liangke_SymCLKG_TKDE}. 

Specifically, when tasks require complex temporal reasoning, LLMs tend to mislead the process and provide an inaccurate result. 
For instance, \textit{``Which country's government leader visited China for the last time in 2015?''}, 
to answer this question, we need to (1) get which countries visited China in 2015 ; (2) filter out the country with the earliest visiting date. 
In step 1, LLMs easily meet hallucinations due to the incomplete training data and the uncertainty of parameterised knowledge.
In step 2, LLMs may lead to the error because of the inaccuracy of the time filtering. Within such temporal reasoning tasks, any misjudgment in the temporal knowledge or errors during the temporal reasoning will culminate in erroneous conclusions.
The problem might stem from the temporal unawareness of LLMs, impeding their ability to track and interpret events over time, particularly in situations requiring subtle and time-sensitive understanding. 

Based on intuitive and empirical analysis, the cause accounting for the problem can be identified from two aspects:
\textit{lack of temporal knowledge} and \textit{lack of complex temporal reasoning}.
And the definition is given as follows: 

\textbf{\textsc{Lack of Temporal Knowledge}}.
LLMs acquire vast knowledge through pre-training on extensive datasets. However, the fixed nature of their parameters after training solidifies their knowledge base, which leads to LLMs' failure in understanding unseen and evolving knowledge.

\textbf{\textsc{Lack of Complex Temporal Reasoning}}.
Owing to the inherent nature of large models that generate outputs based on maximum probability, they are limited in directly conducting complex reasoning. 
Facing the interconnected multi-step temporal reasoning, LLMs might accumulate errors during the process of probabilistic generation.
 
\begin{figure*}[ht]
  \centering
  \includegraphics[width=\linewidth]{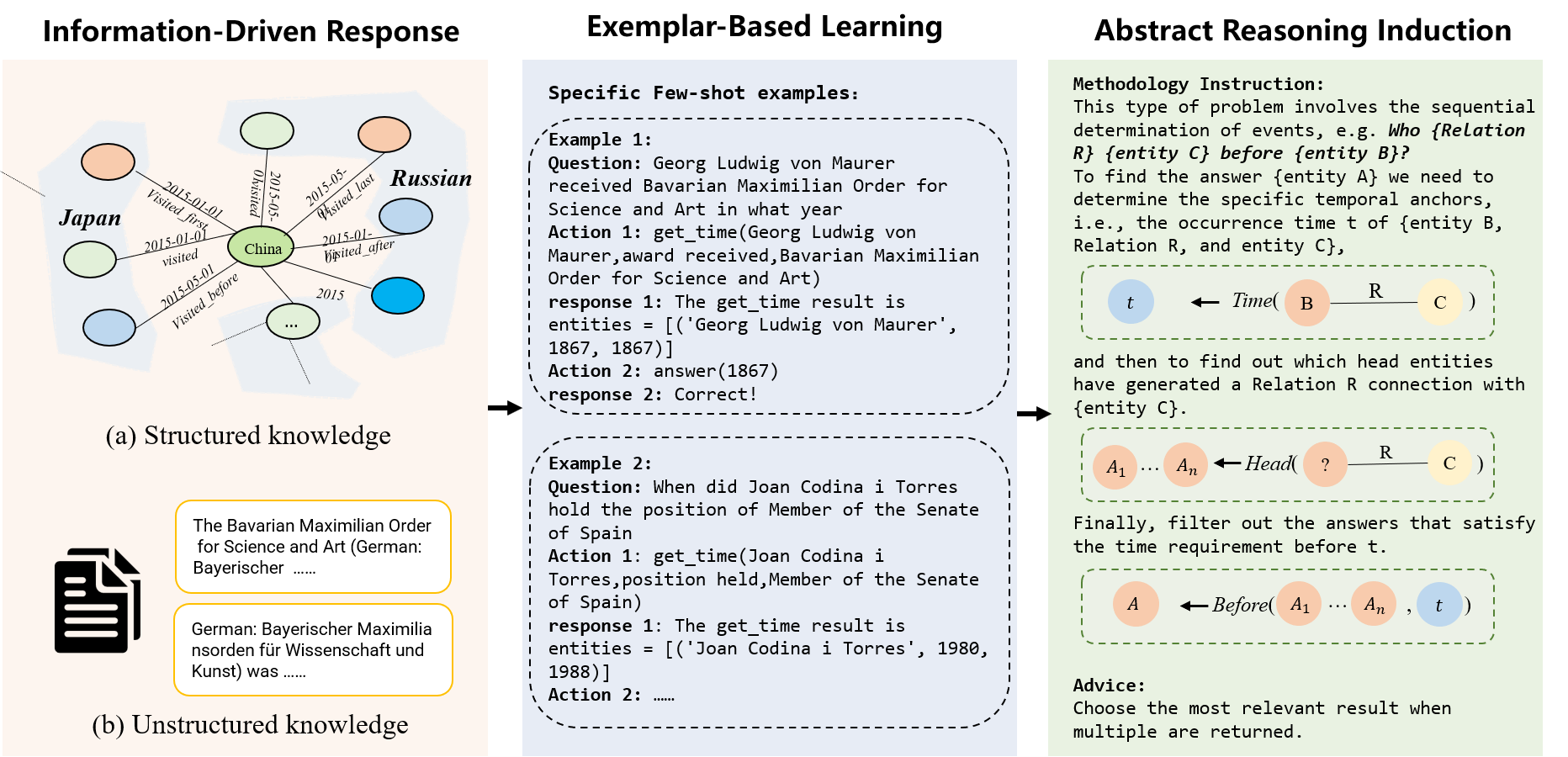}
  \caption{Three levels of information utilisation.  \textit{Information-Driven Response}, which extracts pertinent knowledge to form the basis of answers;  \textit{Exemplar-Based Learning}, offering cases of reasoning for the language model to assimilate and guide current inferences; and \textit{Abstract Reasoning Induction}, providing step-wise abstract methodological guidance to the present question, distinct from concrete knowledge, thereby steering the language model's inference process.
  }
  \label{fig:three_level}
  \vspace{-10pt}
\end{figure*}

Despite the neglect of essences, current studies relatively approach above challenges. 
To augment the LLMs' capacity for understanding unseen and evolving information, researchers incorporate external knowledge to supply contextually relevant information, known as Retrieval Augmented Generation (RAG) ~\cite{DBLP:conf/acl/ZhaoHZSW23,DBLP:journals/corr/abs-2306-04136,DBLP:journals/corr/abs-2307-07697}. 
Although these methods enhance the richness of LLMs' responses, 
the retrieval accuracy and input length limitations might result in irrelevant noises and incomplete reasoning clues, degrading overall performance
~\cite{DBLP:journals/corr/abs-2308-15022,DBLP:journals/corr/abs-2308-08239}.
Furthermore, although tailored examples serve as prompts to guide LLMs~\cite{Dong2023ASF,Min2022RethinkingTR},
they are often inadequate for diverse practical tasks and require substantial efforts in time and human to acquire high-quality examples. 
In conclusion, above approaches fail to provide necessary guidance for ongoing temporal reasoning processes and are susceptible to incorporating extraneous noise, as shown in Figure~\ref{fig:three_level}. 

To overcome the limitation, it is crucial to recognize that LLMs are inherently limited by reliance on passively absorbing training instances. 
Constructivism~\cite{Savery1995ProblemBL,Kirschner2006WhyMG}, deeply embedded in philosophical and psychological schools of thought, contends that knowledge and learning emerge not from mere exposure to external information but through active construction. 
It asserts that learners synthesize new knowledge by building upon their existing understanding and experiences~\cite{lake2023human}.
In this view, learning is an active and ongoing process wherein individuals continuously modify and refine their cognitive frameworks.

Inspired by the principles of constructivism, we try to steer LLMs towards an active and self-initiated learning approach, and propose an Abstract Reasoning Induction (\ourmodel) framework. This will equip LLMs with the capacity for abstract synthesis and personalized knowledge application, enhancing relevance and utility in various contexts.

In details, to handle the lack of temporal knowledge, we transfer the data generation to an active process, consisting of
two stages: \textit{Knowledge-agnostic} and \textit{Knowledge-based}. 
In knowledge-agnostic part, LLMs only need to choose potential steps.
It is only in the knowledge-based part that the corresponding action is executed on the specific knowledge base to obtain the answer.
This procedure offers factual knowledge support to LLMs while minimizing the incorporation of extraneous noisy data. 
On the other hand, to complete LLMs' complex temporal reasoning ability,
\ourmodel actively
engages in proactive and self-directed learning from both correct and incorrect historical reasoning samples. 
This approach enables LLM to summarize and generalize methodologies (i.e. knowledge-agnostic step-by-step instructions) for different types of questions. When similar questions are encountered again, these abstract methods will guide the LLM to perform more efficient multi-step reasoning.
By teaching LLMs to actively construct knowledge and methods, it can significantly boosting their temporal reasoning abilities without the need for further training.

In summary, our contribution is three-fold:
\begin{itemize}
\item Grounded in the principles of constructivism, we offer fresh perspectives for enhancing the reasoning capabilities and task adaptability of LLMs.
\item We present \ourmodel, a novel temporal reasoning framework  that divides the process into two phases:\textit{ Knowledge-agnostic} and \textit{Knowledge-based}.  \ourmodel enables LLMs to learn and construct proactively from historical reasoning samples, fostering a perpetual refinement of LLMs' reasoning abilities.
\item The experimental results demonstrate that, compared to the leading TKGQA models, our approach achieves relative improvements of 29.7\% and 9.27\% respectively on two temporal QA datasets.
\end{itemize}

%% file: chapters/related.tex
\subsection{\tkgqa Models}
Traditional temporal knowledge graph question answering (TKGQA) methodologies fall into two categories. The first, exemplified by \textsf{TEQUILA}~\cite{DBLP:conf/cikm/JiaARSW18}, deconstructs the initial question into sub-questions and temporal constraints, employing standard KGQA models for resolution, followed by a comparative analysis to select the most fitting answer. The second approach, such as \cronkgqa~\cite{DBLP:conf/acl/SaxenaCT20}, seeks to leverage TKG embeddings for semantic similarity assessments in answer determination, featuring a learnable reasoning process independent of handcrafted rules. Despite \cronkgqa's proficiency with simpler inquiries, its performance falters with complex questions necessitating specific temporal inference. \tempqr~\cite{DBLP:journals/corr/abs-2112-05785} addresses this by incorporating temporal scope data and employing the \eae method~\cite{Fvry2020EntitiesAE} to enrich question representation semantically.

However, traditional approaches rely on handcrafted rules or learnable representations, struggling with sophisticated temporal reasoning~\cite{DBLP:journals/kbs/ChenZLLK22}. In contrast, our model, leveraging the power of LLMs, excels in these challenging scenarios, showcasing superior adaptability and reasoning capabilities.

\subsection{LLM Reasoning with External Information}
Addressing hallucinations in generative models presents a compelling challenge, with one promising solution being the augmentation of LLMs with external knowledge~\cite{DBLP:journals/corr/abs-2302-07842}. Integration with an external knowledge base has become a prevalent strategy in question-answering and conversational tasks~\cite{DBLP:journals/corr/abs-2302-12813}. There are mainly two approaches: explicit and implicit knowledge injection~\cite{DBLP:journals/corr/abs-2306-11489}. Explicit injection involves directly supplying LLMs with pertinent knowledge via prompts. For instance, \textsf{KAPING}~\cite{DBLP:journals/corr/abs-2306-04136} retrieves facts relevant to a query from a knowledge graph and appends these to the query as a prompt for the LLM, while \textsf{CoK}~\cite{Li2023ChainofKnowledgeGL} first evaluates answer credibility and, if necessary, uses the LLM to decompose the question and generate various SPARQL queries to extract information from external knowledge bases. 
\textsf{ChatKBQA}~\cite{DBLP:journals/corr/abs-2310-08975}  finetunes LLMs on KG structure to generate logical queries, which can be executed on KGs to obtain answers.
\textsf{Symbol-LLM}~\cite{DBLP:journals/corr/abs-2311-09278}  propose a dual-stage fine-tuning framework to integrates symbolic knowledge into LLMs, enhancing their reasoning capabilities.
\textsf{ToG}~\cite{DBLP:journals/corr/abs-2307-07697} treats the LLM as an agent to interactively explore related entities and relations on KGs and perform reasoning based on the retrieved knowledge.
Implicit injection, on the other hand, subtly steers the LLM by incorporating knowledge semantic embeddings during reasoning or in the decoding process. \textsf{KID}~\cite{DBLP:conf/iclr/LiuZGGGVSA22} represents a novel decoding algorithm for generative LMs that dynamically infuses external knowledge at each step of LM decoding, and \textsf{KPE}~\cite{DBLP:conf/acl/ZhaoHZSW23} introduces a trainable parameter-sharing adapter to a parameter-freezing PLM for knowledge integration. Additionally, the incorporation of knowledge from other modalities such as visual information~\cite{li2023towards,li2023lmeye} has also become a method of knowledge introduction.

While the integration of knowledge into LLMs can mitigate issues of hallucinations, it is not without challenges. Explicit knowledge injection often struggles to acquire high-quality, relevant information, and is constrained by the finite-length contexts. Implicit injection typically necessitates fine-tuning of parameters, which can be prohibitively costly. We address these limitations by dividing temporal knowledge reasoning into two distinct components: knowledge-related and knowledge-agnostic. This approach achieves a clear separation between knowledge and reasoning, thereby circumventing the aforementioned constraints.

\subsection{LLM Reasoning with Memories}

Memory plays a pivotal role in human intelligence~\cite{Atkinson1968HumanMA}. Given that LLMs inherently lack long-term memory and their short-term memory is constrained by the scope of their context window, numerous studies have embarked on the journey to equip LLMs with memory capabilities~\cite{DBLP:journals/corr/abs-2308-03188,DBLP:journals/corr/abs-2305-10250}.
Instead of the conventional approach where accumulated conversations are retrieved directly, \textsf{MemoChat}~\cite{DBLP:journals/corr/abs-2308-08239}  innovatively constructs and updates a structured, instant memo that categorizes past dialogues. Conversations are then fetched based on their specific topics and summaries.
\textsf{Reflexion}~\cite{DBLP:journals/corr/abs-2303-11366} exploits a working memory to store experiences for a dedicated task to improve the performance of the agent through several trials.
However, the histories stored in working memory cannot benefit the episode for different task goals.
\textsf{MemPrompt}~\cite{DBLP:conf/emnlp/MadaanTCY22} designs a persistent memory to store human feedback to remind the chatbot of the conversational information and improve it continuously. 
\textsf{RLEM}~\cite{zhang2023large} adopts a persistent environment-grounded experience memory to store the experiences and assist in future decision-making even for different task goal.
\textsf{Thought Propagation}~\cite{DBLP:journals/corr/abs-2310-03965} emphasizes the ability to explore and apply insights from analogous solutions. By delving into and utilizing solutions from problems related to the given issue, it improves performance and accuracy across various tasks.
 
However, current memory-enhanced methods are limited to passively received historical information, overlooking the active construction of abstract knowledge based on previous experience. Starting from constructivism,  we apply the proposed method to provide large models with an active and continuous learning process, offering knowledge that is abstract and generalized.

%% file: chapters/method.tex
\begin{figure*}[ht]
  \centering
  \includegraphics[width=\linewidth]{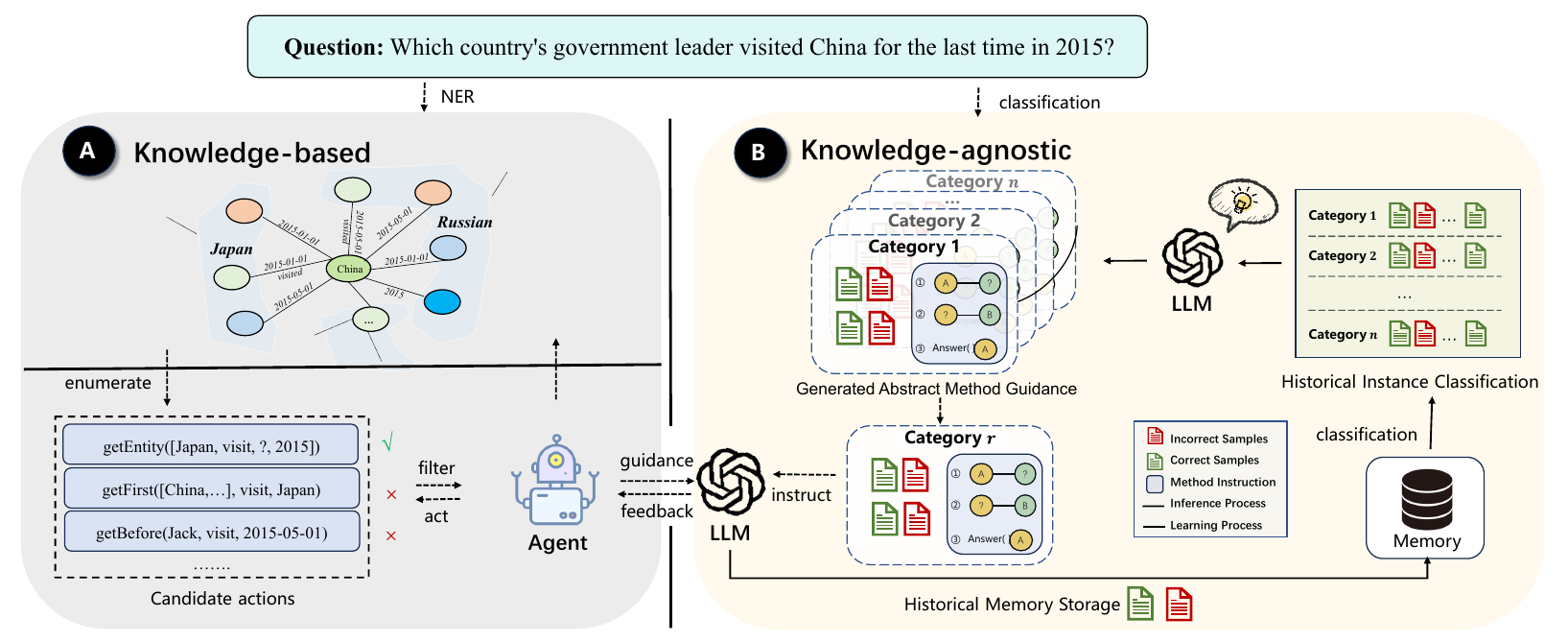}
  \caption{Model architecture of \ourmodel. 
   Our framework divides temporal reasoning into two distinct phases: \textit{Knowledge-agnostic} and \textit{Knowledge-based}. This division aims to reduce instances of hallucinations and improve LLM's capacity for integrating abstract methodologies derived from historical experience. See the detailed instructions in Appendix~\ref{sec:instruct}.
  }
  \label{fig:model}
\end{figure*}

\begin{algorithm}[ht]
\label{ari_algorithm}
\caption{Abstract Reasoning Induction}
\begin{algorithmic}[1]
\REQUIRE Temporal knowledge graph $\mathcal{K}$, question $q$, historical memory $H_q$, abstract methodology instruction set $M_{C}$
\ENSURE Answer to the question $q$ 
\STATE $M_{C} \leftarrow LLM(H_q)$    \hfill \eqref{eq:learning}
\STATE Initialize subject entity $e_h$ from $q$ 
\STATE Find 1-hop subgraph $G_{e_h}$ of $e_h$ in $\mathcal{K}$ \hfill \eqref{eq:subgraph}
\STATE Enumerate initial candidate actions $P_0$ from $G_{e_h}$ \hfill \eqref{eq:action_generate}
\WHILE{$LLM(M_{C^*},P^{'}_{t_i}) \ne answer(a)$}
    \STATE Filter candidate actions $P_{t_i}$ to get $P^{'}_{t_i}$ 
    \hfill \eqref{eq:action:filter}
    \STATE $ C^*_t \leftarrow $ findKmeansCluster(q) \hfill \eqref{eq:find_k}
    \STATE  $a^*_i =  LLM(M_{C^*},q,P^{'}_{t_i})$ \hfill \eqref{eq:answer}
    \STATE Execute selected action $a^*_i$ and update current environment 
    \STATE Regenerate candidate actions for the next step \hfill \eqref{eq:action_generate}
    \IF{$t_i \geq t_{max}$} 
        \STATE Break
    \ENDIF
\ENDWHILE

\STATE Execute final action $a^*_i$ to abtain the answer
\STATE Add current process to $H_q$
\RETURN Answer derived from the reasoning process
\end{algorithmic}
\end{algorithm}

\subsection{Task Definition}

Given a Temporal Knowledge Graph (TKG) $\mathcal{K}$ and a natural language question $q$, 
TKGQA aims to extract an entity $s/o \in \mathcal{E}$ or a timestamp $\tau \in \mathcal{T}$ that correctly answers the question $q$. For instance, for the question \textit{`Which country's government leader visited China for the last time in 2015?'} Based on the event information contained in the TKG, we can get the answer to the question is the KG entity \textit{Vietnam}.

\subsection{ARI Framework}
\label{sec:overall}
The constructivist perspective posits that knowledge does not merely encapsulate universal laws but must be contextually reconstructed for specific situations. This view emphasizes that understanding is a construct developed by the learner, uniquely shaped by their experiential background and dependent on their learning trajectory in a particular context~\cite{Kirschner2006WhyMG, Savery1995ProblemBL}. In line with this philosophy, we introduce \ourmodel framework.
We overview the framework in Figure~\ref{fig:model}.
Diverging from previous research that directly feeds knowledge into LLMs
, we divide temporal knowledge reasoning into two parts: \textit{knowledge-agnostic} and\textit{ knowledge-based}.

The knowledge-based module extracts relevant TKG subgraphs based on the given question, generating all feasible fine-grained actions through traversal  
(cf. \S~\ref{sec:knowledge-based}). 
This module encompasses all operations and interactions with knowledge, freeing the LLM from the vast, noisy specific information to focus on reasoning. 
This design approach allows us to build intricate knowledge queries by combining fine-grained atomic operations, while effortlessly adapting to various knowledge bases (see the detailed operations in
Appendix~\ref{sec:actions}). In the knowledge-agnostic module, the LLM performs high-level strategic decisions and candidate action selection. We have innovated mechanisms that allow LLMs to internalize lessons from past decisions, forming generalized abstract methodologies (solid line process in Figure~\ref{fig:model}). This foundation empowers the LLM to proactively develop abstract methodological guidelines for various question types, enhancing its ability to reason on new questions efficiently (cf. \S\ref{sec:knowledge-agnostic}).


For the inference process, \ourmodel first categorizes the question and selects the most suitable abstract methodological guidance. According to these guidelines, the LLM interacts multiple turns with the knowledge-agnostic module to reason and gradually solve the problem, as shown in the dashed line process in Figure~\ref{fig:model}. We also present a specific reasoning example in Figure~\ref{fig:ill} of Appendix~\ref{sec:instruct}.

\subsection{Knowledge-based Interaction}
\label{sec:knowledge-based}

In the knowledge-based interaction part, we frame complex temporal knowledge reasoning challenges as multi-step inferentce tasks~\cite{DBLP:conf/coling/Gu022,DBLP:conf/acl/Gu0023}. At the beginning of each step, we employ an filtering mechanism to engage with the TKG and the current question. This interaction produces a set of feasible candidate actions for each step. The LLM then selects the most suitable action from these candidates. Following this selection, the model interacts with the TKG, updating the initial state for the next step in a recursive process.

\myparagraph{Candidate Action Enumeration} Specifically, given a complex temporal question $q$ and a TKG $\mathcal{K}:=(\mathcal{E}, \mathcal{R}, \mathcal{T}, \mathcal{F})$, where $\mathcal{E},\mathcal{R}, \mathcal{T}$ denote entities, relations, and timestamps, respectively. 
Starting from the subject entity $e_h$ of $q$, we first find the 1-hop subgraph of $e_h$ in $\mathcal{K}$. Let $N_{e_h}$ be the set of nodes in the 1-hop (undirected) neighborhood of $e_h$ in the TKG, and $R_{e_h}$ be the corresponding edge. 
\begin{equation}
    \label{eq:subgraph}
    \begin{split}
        G_{e_h} = \{(e,r)|e \in N_{e_h}, r \in R_{e_h}\},
    \end{split}
\end{equation}
where $G_{e_h}$ is the corresponding 1-hop subgraph of $e_h$. For each edge $r \in G_{e_h}$,
our agent will strictly follow finely-grained atomic operations in Appendix~\ref{sec:actions}, traversing and replacing the relations and entities present in the current $G_{e_h}$ to construct the set of candidate actions $P_0$, where the subscript 0 denotes the candidate actions for the initial step,
\begin{equation}
    \label{eq:action_generate}
    \begin{split}
        P_0 = \{Enum(action,e,r)|e,r \in G_{e_h}\}.
    \end{split}
\end{equation}
\myparagraph{Candidate Action Filtration} However, due to the continual occurrence and updating of temporal events, even the scale of a 1-hop subgraph can be vast. This results in an excessively large set of generated candidate actions, which can significantly impede the judgment of LLMs (cf. \S~section~\ref{sec:results}). Consequently, we propose a filtration process for candidate actions, retaining only those that are correct, feasible, and semantically relevant.

Specifically, for each action $a$ within set $P_0$, we execute the corresponding function on the TKG. If the function returns a non-empty value, the action is considered correct and feasible; otherwise, it is discarded. Among all remaining actions, we retain the top-$K$ actions based on the calculation of their semantic similarity to the question $q$,
\begin{equation}
    \label{eq:action:filter}
    \begin{split}
        P^{'}_{0} = \{a | \text{exec}(a) \neq \emptyset \wedge a \in \text{Top-}K(P_0,q) \}.
    \end{split}
\end{equation}
Based on the LLM's decision, the agent executes the corresponding action, thereby updating the current environment. Subsequently, it regenerates the next set of candidates $P_1$ based on the newly identified subject entities, repeating this process until a termination command is received.

\subsection{Knowledge-agnostic Reasoning}
\label{sec:knowledge-agnostic}

The knowledge-agnostic module enables LLMs to distill and apply abstract methodologies from historical reasoning examples, enabling adaptation to diverse questions. This approach fosters a general methodology, applicable across various domains and to a wide range of knowledge-independent inquiries, enhancing LLMs' versatility.

\myparagraph{Historical Memory Storage and Learning }
In the LLM's reasoning process, we meticulously document the current state at each step \( t_i \), encompassing the current temporal question $q$, the set of candidate actions $P_t$, and the LLM's decision $a_t$. The aggregate of all stepwise states for a given question forms the historical decision set $H_q$,
\begin{equation}
    \label{eq:learning}
    \begin{split}
        H_q = \{ (q, t_i, P_{t_i}, a_{t_i}) \mid i \in T \},
    \end{split}
\end{equation}
where $T$ is the set of all steps in the process. 

Temporal reasoning is often multi-step and complex, yet the types of reasoning involved tend to be consistent, with similar questions requiring similar inference steps. Therefore, once the LLM conducts reasoning and accumulates a series of historical inferential steps, we employ unsupervised clustering K-means~\cite{MacQueen1967SomeMF} to categorize these historical steps into distinct clusters $C_H$. 

Consequently, after the LLM engages in reasoning and compiles historical reasoning steps, these are subjected to unsupervised clustering to form distinct clusters. Each cluster encapsulates a mix of both accurate and erroneous reasoning processes. Subsequently, we enable the LLM to actively learn from these specific historical instances within each cluster and distill abstract methodologies that are independent of domain-specific knowledge.

\myparagraph{LLM Decision with Abstract Reasoning} When addressing new inference challenges, we initiate the process by identifying the historical reasoning cluster most closely aligned with the new question. We then extract its abstract methodologies to guide the LLM in its reasoning for the current question. 

Specifically, for a given question $q$, we calculate the similarity score $S(C_i,q)$ with each historical reasoning cluster $C_i$. We then retrieve the abstract method instruction $M_{C^*}$ from the cluster that yields the maximum $S$.
Let \( \{C_1, C_2, \ldots, C_n\} \) be the set of historical reasoning clusters. For a given question \( q \), we calculate the similarity score \( S(C_i, q) \) for each cluster \( C_i \), and select the cluster $C^*$ with the highest similarity score to the query $q$,
\begin{equation}
    \label{eq:find_k}
    \begin{split}
    C^* = \underset{C_i}{\mathrm{argmax}}\ S(C_i, q),
    \end{split}
\end{equation}
The abstract method instruction \( M_{C^*} \) is then selected from the cluster \( C^* \) such that:
\begin{equation}
    \label{eq:answer}
    \begin{split}
    a^*_i = LLM(M_{C^*},q,P_i),
    \end{split}
\end{equation}
where \( M_{C^*} \) is the abstract method instruction of \( C^* \), and $a^*_i$ is the final output of LLM. The reasoning sequence concludes when the LLM outputs a termination action or when the length of reasoning steps exceeds the predetermined maximum threshold.

%% file: chapters/experiment.tex
\begin{table*}[ht]
    \resizebox{\textwidth}{!}{%
    \begin{tabular}{c|c|cc|cc|ccccc}
    \hline
    \multicolumn{1}{c|}{\multirow{3}{*}{\textbf{Model}}} & \multicolumn{5}{c|}{\textbf{\ourq}} & \multicolumn{5}{c}{\textbf{\cronq}} \\ 
    \cline{2-11} 
    \multicolumn{1}{c|}{} & \multicolumn{1}{c|}{\multirow{2}{*}{\textbf{Overall}}} & \multicolumn{2}{c|}{\textbf{Question Type}} & \multicolumn{2}{c|}{\textbf{Answer Type}} & \multicolumn{1}{c|}{\multirow{2}{*}{\textbf{Overall}}} & \multicolumn{2}{c|}{\textbf{Question Type}} & \multicolumn{2}{c}{\textbf{Answer Type}} \\ 
    \cline{3-6} \cline{8-11} 
    \multicolumn{1}{c|}{} & \multicolumn{1}{c|}{} & \multicolumn{1}{c|}{\textbf{Simple}} & \multicolumn{1}{c|}{\textbf{Complex}} & \multicolumn{1}{c|}{\textbf{Entity}} & \multicolumn{1}{c|}{\textbf{Time}} & \multicolumn{1}{c|}{} & \multicolumn{1}{c|}{\textbf{Simple}} & \multicolumn{1}{c|}{\textbf{Complex}} & \multicolumn{1}{c|}{\textbf{Entity}} & \multicolumn{1}{c}{\textbf{Time}} \\ 
    \hline
    \textsf{BERT} & 0.083 & 0.092 & 0.061 & 0.101 & 0.040 & 0.243 & 0.249 & 0.239 & 0.277 & 0.179 \\
    \textsf{ALBERT} & 0.108 & 0.116 & 0.086 & 0.139 & 0.032 & 0.248 & 0.255 & 0.235 & 0.279 & 0.177 \\
    \textsf{EmbedKGQA} & 0.206 & 0.235 & 0.134 & 0.290 & 0.001 & 0.288 & 0.290 & 0.286 & 0.411 & 0.057 \\
    \cronkgqa & 0.279 & 0.134 & 0.134 & 0.328 & 0.156 & 0.647 & 0.987 & 0.392 & 0.699 & 0.549 \\ 
    \multitq & 0.293 & 0.347 & 0.159 & 0.349 & 0.157 & - & - & - & - & - \\
    \chatgpt & 0.102 & 0.147 & 0.077 & 0.137 & 0.002 & 0.249 & 0.250 & 0.247 & 0.246 & 0.253 \\ 
    \rag & 0.185 & 0.200 & 0.160 & 0.230 & 0.07 & 0.490 & 0.460 & 0.518 & 0.470 & 0.520 \\ 
    \chainofthought\textsubscript{KB} & 0.240 & 0.440 & 0.120 & 0.220 & 0.320 & 0.640 & 0.690 & 0.610 & 0.620 & 0.660 \\ 
    \react\textsubscript{KB} & 0.310 & 0.635 & 0.136 & 0.313 & 0.300 & 0.685 & 0.835 & 0.525 & 0.650 & 0.755 \\ 

    \hline
    \textbf{\ourmodel} & \textbf{0.380$^{**}$} & \textbf{0.680$^{**}$} & \textbf{0.210$^{**}$} & \textbf{0.394$^{**}$} & \textbf{0.344$^{**}$} & \textbf{0.707$^{**}$} & 0.860 & \textbf{0.570$^{**}$} & \textbf{0.660$^{*}$} & \textbf{0.800$^{*}$} \\
    \hline
    \end{tabular}%
    }
    \caption{Performance of baselines and our methods on the \ourq and \cronq. ${^*}(p \leq 0.05)$ and $^{**} (p \leq 0.005) $ indicate paired t-test of \ourmodel versus the best baseline.}
    \label{tab:experiment}
\end{table*}

\section{Experiment}

\subsection{Implementation Details}
We use \texttt{gpt-3.5-turbo-0613} as our LLM (More experiments and analyses of other LLMs can be found in Section~\ref{sec:other_llm}). We configure the LLM to access and investigate a corpus of 200 historical reasoning samples, with the maximum length of reasoning path set to 5, and the number of historical path categories fixed at 10. Due to the vast size of the test set, which comprises more than 50,000 question-answer pairs, we employ a stratified sampling approach for evaluation, extracting a subset of 200 questions from the test set for each iteration.
In our evaluation, we compare several baseline methods, including the traditional TKGQA models and LLM-based models 
(see Appendix~\ref{sec:tkgqa_baselines} 
 and \ref{sec:datasets} for more details).

\subsection{Overall Results}
\label{sec:results}
Table~\ref{tab:experiment} presents the comparative results of \ourmodel against other baselines on \ourq. 
\paragraph{LLM only Performance.}\chatgpt's performance on two datasets reveal a significant shortcoming in its application to temporal knowledge reasoning, even when all the knowledge required for the questions is within the scope of its training data prior to 2021. This deficiency is particularly pronounced when compared to traditional TKGQA methods, suggesting that the parameterised knowledge acquired by LLMs is not seamlessly transferred to temporal reasoning tasks.
A stark contrast in performance is observed between the \ourq and \cronq datasets, the latter benefiting from \chatgpt's extensive training on \texttt{WikiData}\cite{DBLP:journals/cacm/VrandecicK14}. This discrepancy points to a significant challenge: \ourq's reliance on the \texttt{ICEWS}\cite{DVN/28075_2015}, which encompasses more niche and frequent events, poses a difficulty for LLMs due to their limited exposure to such data during training. This contrast elucidates the inherent limitations of LLMs in processing temporal knowledge.


\myparagraph{Reasoning with External Knowledge} Incorporating additional knowledge graph data into LLMs, as seen with \rag, considerably enhances their performance in knowledge-intensive QA tasks. \rag outperforms \chatgpt by 81\% and 96\% across two datasets, respectively. Nevertheless, it still falls short of leading TKGQA models. This gap can be attributed to two main factors. First, the vast and complex nature of temporal information presents a challenge. A single question may involve thousands of related events, which cannot be accurately incorporated through prompts alone, leading to insufficient background information for reasoning. Second, the retrieved external knowledge often contains redundant or irrelevant information, which can further mislead the model's inference process.

\myparagraph{Reasoning with Exemplar Guidance}The \chainofthought\textsubscript{KB} approach, despite providing step-by-step reasoning guidance and knowledge-based interaction, does not reach the efficacy levels of \ourmodel. 
This discrepancy stems from the exemplar-based learning method's dependence on specific instances, which can introduce extraneous knowledge and detract LLMs from focusing on reasoning methodologies, leading to inaccurate conclusions. Additionally, its static examples fail to provide the customized guidance necessary for diverse reasoning questions. 
Similarly, our investigation also extends to the augmented knowledge-agnostic module in \react, designed to mitigate the generation of infeasible actions. Despite this enhancement, a performance gap persists when compared to \ourmodel, underscoring similar limitations identified in the \chainofthought\textsubscript{KB} approach. Despite \react\textsubscript{KB}'s interaction with the environment, it still lacks customized abstract guidance for the current reasoning question.

\ourmodel significantly outperforms current state-of-the-art TKGQA models, achieving a relative improvement of 29.7\% on the \ourq dataset and a 9.27\% increase in performance on the \cronq dataset. 
These substantial gains can be attributed to the knowledge adaptability and the abstract methodology instruction mechanism, which empower LLMs to make advanced decisions. By leveraging abstract methodologies, LLMs can select optimal temporal reasoning steps without engaging with the specifics of the underlying knowledge.

\begin{figure}[ht]
  \centering
  \includegraphics[width=\linewidth]{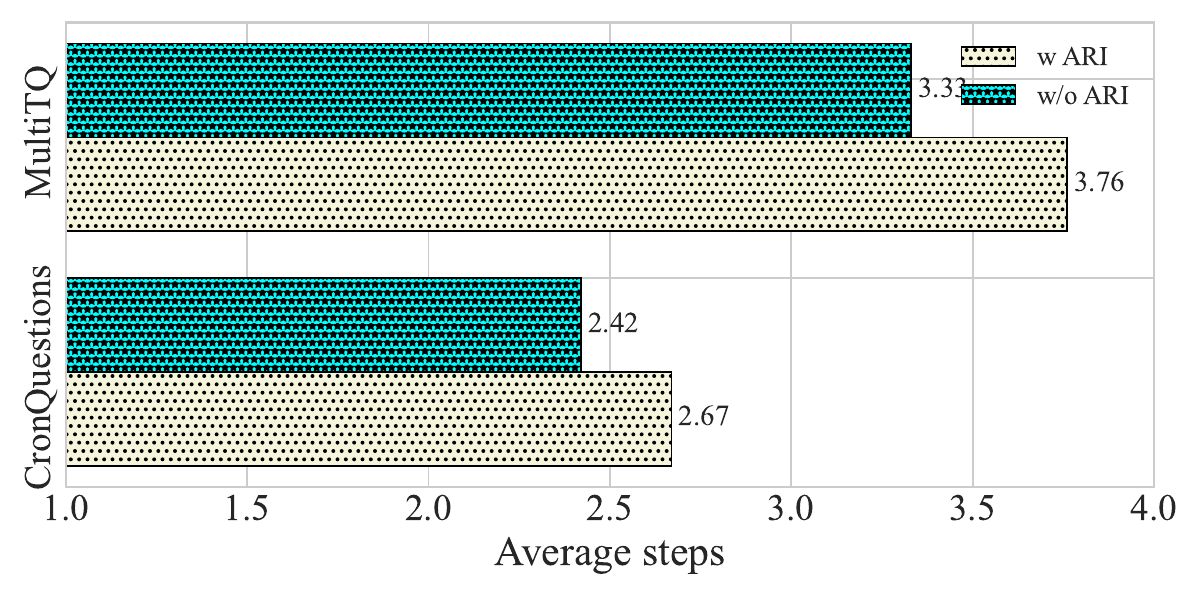}
  \caption{Comparison of average reasoning steps of \ourmodel on \ourq.
  }
  \label{fig:length}
  \vspace{-10pt}
\end{figure}

\myparagraph{Comparison of Reasoning Efficiency}
To validate the effectiveness of abstract instruction, we conduct an evaluation of reasoning efficiency. On the test set, with all other components of the model remaining constant, we remove the abstract instruction and record the average number of steps taken for reasoning. Compared with the \ourmodel, we observe that under the guidance of abstract methodologies, LLMs not only improve in reasoning accuracy but also reduce their average number of reasoning steps by 11.4\% on \ourq and 9.3\% on \cronq. This underscores that the guidance provided by abstract methodologies can significantly enhance the efficiency of LLMs in temporal reasoning tasks.

\begin{table}[t]\centering
\resizebox{\linewidth}{!}{
\begin{tabular}{lcc}\toprule
\multirow{2}{*}{\textbf{Model}} &\multicolumn{2}{c}{\textbf{Accuracy (\%)}} \\
\cmidrule(lr){2-3}
&\ourq &\cronq \\
\cmidrule[\heavyrulewidth]{1-3}
\textbf{\ourmodel}         &38.0    &70.7   \\           
\cmidrule{1-3}
w/o Abstract Guidance        &30.5    &67.1   \\
w/o History Cluster  &34.5   &68.9   \\
w/o Action Filter       &33.1    &66.5 \\
w/o Incorrect Examples	 &36.5    &69.2 \\
\bottomrule
\end{tabular}
}
\caption{Ablation results of \ourmodel.}
\label{tab:ablation_study}
\end{table}

\subsection{Ablation Study}
To evaluate the efficacy of the individual components of the model, we conducted ablation studies.

Initially, we remove the abstract guidance component, requiring the LLM to rely on its own understanding of the questions without the aid of historical information. This result in significant performance drops on both datasets, with a 19.7\% decrease on \ourq and a 3.7\% decrease on \cronq. This suggests that distilled abstract guidance plays a substantial role in supporting the model's reasoning capabilities.

To further assess the impact of abstract guidance, we eliminate the clustering module, thus deriving a universal abstract guidance from all historical reasoning processes without categorization based on question type. The model performance dropped by 9.2\% on \ourq and 2.5\% on \cronq, indicating that a singular abstract methodology is insufficient for guiding various types of questions and that targeted abstract methodological guidance is more effective. In Appendix~\ref{sec:Cluster}, we illustrate the impact of varying cluster quantity on the model's final reasoning performance.

To verify the role of incorrect samples in \ourmodel, we remove incorrect examples from the process of generating abstract methods, providing only correct examples as guidance. As evident from the results, the removal of incorrect examples leads to a decrease in the quality of abstract guidance, resulting in a performance drop. 
By encountering and learning these incorrect examples, LLMs become more adept at avoiding similar pitfalls in subsequent reasoning tasks. This also aligns with our intuition and has been validated in prior studies~\cite{DBLP:conf/emnlp/WangL23,DBLP:conf/emnlp/YangLL23,DBLP:journals/corr/abs-2310-20689}.

Lastly, we remove the action selection module, allowing the LLM to choose from all generated actions without filtering. This led to a decrease in performance on both datasets by 12.8\% and 5.9\%, underscoring that unfiltered actions result in an excessive number of options, including irrelevant ones, which hinders the LLM's reasoning and complicates the decision-making process.

%% file: chapters/conclusion.tex
This study, anchored in the principles of constructivism, critically examines the shortcomings of LLMs in addressing complex temporal reasoning challenges and proposes an innovative approach to augment their reasoning capabilities. Through the integration of a knowledge adaptability framework and abstract methodological guidance, we have shown that LLMs can attain more precise and efficient reasoning in complex temporal scenarios, effectively overcoming their constraints in processing and interpreting time-sensitive knowledge. 

\myparagraph{Limitations}
While \ourmodel demonstrates impressive results, it also presents several limitations for ongoing refinement.
Firstly, The efficacy of generating abstract guidance heavily relies on the capabilities of LLMs. Smaller-scale LLMs may struggle to produce high-quality abstract guidance, thus potentially restricting their application. 
Secondly, the \ourmodel framework depends on multi-step reasoning to arrive at final answers, a process moderately influenced by the LLM's reasoning efficiency, which extends the duration of inference.
Finally, our method is primarily concentrated on complex temporal reasoning, with its effectiveness in other reasoning domains remaining to be examined. 
Future research should aim to refine these methods to make them more adaptable to various models and problem domains, enhance the balance between reasoning efficiency and depth, and expand their scope to include a broader range of reasoning tasks.

%% file: chapters/acknowledgement.tex
The authors would like to thank the anonymous reviewers for their insightful and constructive comments, which greatly contributed to improving the quality of the paper. 
This work was partially supported by National Key R\&D Program of China (No. 2022YFB3103600),
NSFC (Nos. U23A20296, 62272469), and The Science and Technology Innovation Program of Hunan Province (No. 2023RC1007).

%% file: chapters/appendix.tex
\subsection{More Details about Baseline Methods}
\label{sec:tkgqa_baselines}
In our evaluation, we compared several baseline methods. 
\begin{itemize}
    \item \textbf{Pre-trained LMs}: To evaluate \textsf{BERT}~\cite{DBLP:conf/naacl/DevlinCLT19} and \textsf{ALBERT}~\cite{DBLP:conf/iclr/LanCGGSS20}, we generate their LM-based question embedding and concatenate it with the entity and time embeddings, followed by a learnable projection. The resulted embedding is scored against all entities and timestamps via dot-product.
    \item \textbf{\embed}~\cite{DBLP:conf/acl/SaxenaTT20} is designed with static KGs. To deal with multiple temporal granularities, 
    timestamps are ignored during pre-training and random time embeddings are used. 
    \item \textbf{\cronkgqa}~\cite{DBLP:conf/acl/SaxenaCT20} is designed for single temporal granularity. To deal with multiple granularities, 
    time embeddings at the year/month granularity are drawn at random from corresponding day embeddings.
    \item \textbf{\multitq}~\cite{chen-etal-2023-multi} is designed for multi-granularity temporal granularity with a transformer-based time aggregation module.
    \item \textbf{\chatgpt~\footnote{https://chat.openai.com/}}. We use \chatgpt to provide direct answers to the questions.
    \item \textbf{\rag}. To validate the performance of the LLM in the presence of relevant background knowledge, we extracted relevant quaternions (up to 20) from the TKG based on the entity and time information appearing in the question, and put them in the prompt for \chatgpt to answer as a retrieval-enhanced way of comparison.
    \item \textbf{\react\textsubscript{KB}}:To address the applicability of \react~\cite{DBLP:conf/iclr/YaoZYDSN023} to our task, we designed a variant of \react by integrating our knowledge-agnostic module, which generates all feasible actions (using the same atomic action templates as \ourmodel). LLMs were then prompted to select one action from the available options. Parameter settings remained consistent with our \ourmodel approach, including the use of the same Named Entity Linking (NEL) method and reasoning length.
    
    \item \textbf{\chainofthought\textsubscript{KB}}:We introduce the \chainofthought\textsubscript{KB} method, which integrates a knowledge-based module into the Chain-of-Thought (CoT)~\cite{DBLP:conf/nips/Wei0SBIXCLZ22} framework. This allows the LLM to interact with KG under the guidance of examples, thereby obtaining the final answer more effectively.
    We manually constructing 9 specific instance examples with detailed reasoning steps to guide the LLM, while keeping other settings unchanged. 
\end{itemize}

\begin{table}
\centering \small
\begin{tabular}{l|c|ccc} 
\toprule
\multicolumn{2}{l|}{}                    & \textbf{Train} & \textbf{Dev} & \textbf{Test}  \\ 
\hline
\multirow{3}{*}{Single}   & Equal        & 135,890        & 18,983       & 17,311         \\
                          & Before/After & 75,340         & 11,655       & 11,073         \\
                          & First/Last   & 72,252         & 11,097       & 10,480         \\ 
\hline
\multirow{3}{*}{Multiple} & Equal~Multi  & 16,893         & 3,213        & 3,207          \\
                          & After First  & 43,305         & 6,499        & 6,266          \\
                          & Before Last  & 43,107         & 6,532        & 6,247          \\ 
\hline
\multicolumn{2}{c|}{\textbf{Total}}      & 386,787        & 587,979       & 54,584         \\
\bottomrule
\end{tabular}
\caption{Statistics of question categories in \ourq.}
\label{table:Statistics_mulititq}
\end{table}

\begin{table}
\centering \small
\begin{tabular}{l|c|ccc} 
\toprule
\multicolumn{2}{l|}{}                    & \textbf{Train} & \textbf{Dev} & \textbf{Test}  \\ 
\hline
\multirow{2}{*}{Simple}   & Simple Entity        & 90,651        & 7,745       & 7,812        \\
                          & Simple Time & 61,471         & 5,197       & 5,046         \\
\hline
\multirow{3}{*}{Complex} & Time Join  & 55,453         & 3,878        & 3,832          \\
                          & First/Last  & 118,556         & 11,198        & 11,159          \\
                          & Before/After  & 23,869         &1,928        & 2,151          \\ 
\hline
\multicolumn{2}{c|}{\textbf{Total}}      & 350,000        & 30,000       & 30,000         \\
\bottomrule
\end{tabular}
\caption{Statistics of question categories in \cronq.}
\label{table:Statistics_cronq}
\end{table}

\subsection{Datasets Statistics}
\label{sec:datasets}
\paragraph{\cronq}~\cite{Saxena2021QuestionAO} is a dataset for temporal knowledge graph question answering. The entities and times present in the questions are annotated. \cronq has four question types, including both simple and complex temporal questions. 

\paragraph{\ourq}~\cite{chen-etal-2023-multi} is a complex temporal question answering dataset with multi-granularity temporal information. Compared to existing datasets, \ourq features in a few advantages, including large scale, ample relations and multiple temporal granularity, which hence better reflects real-world scenarios.

We summarize the number of questions in \ourq across different types in Table~\ref{table:Statistics_mulititq} and Table~\ref{table:Statistics_cronq}. In Table~\ref{tab:examples}, we present sample questions from \ourq as per question type, time granularity and answer type.

\begin{table}[ht]
\centering \small
\begin{tabular}{lc}
\toprule
\textbf{Property} & \textbf{Sample Question}                                                                                                        \\ \hline
\multicolumn{2}{l}{\textbf{By question type}}                                                                                             \\ \hline
Equal             & \textit{\begin{tabular}[c]{@{}l@{}}Which country provided humanitarian \\ aid to Sudan in 2007?\end{tabular}}   \\ \hline
Before/After & \textit{\begin{tabular}[c]{@{}l@{}}Who commended the Military of Mali \\ before the Armed Rebel of Mali did?\end{tabular}}          \\ \hline
First/Last   & \textit{\begin{tabular}[c]{@{}l@{}}When did the Militant of Taliban first \\ commend the Government of Pakistan?\end{tabular}}      \\ \hline
Equal Multi       & \textit{\begin{tabular}[c]{@{}l@{}}In 2012, who last did Barack Obama \\ appeal for?\end{tabular}}                              \\ \hline
Before Last       & \textit{\begin{tabular}[c]{@{}l@{}}Who was threatened by Benjamin \\ Netanyahu last before Middle East?\end{tabular}}            \\ \hline
After First  & \textit{\begin{tabular}[c]{@{}l@{}}Who first wanted to negotiate with Evo\\ Morales after the Citizen of Brazil did?\end{tabular}} \\ \hline
\multicolumn{2}{l}{\textbf{By time granularity}}                                                                                          \\ \hline
Year              & \textit{\begin{tabular}[c]{@{}l@{}}Who first made Abu Sayyaf suffer from \\  conventional military forces In 2015?\end{tabular}} \\ \hline
Month             & \textit{\begin{tabular}[c]{@{}l@{}}In Dec, 2008, who would wish to \\ negotiate with the Senate of Romania?\end{tabular}}        \\ \hline
Day               & \textit{\begin{tabular}[c]{@{}l@{}}In Jul 21st, 2011, who criticized the\\ Media of Ecuador?\end{tabular}}                      \\ \hline
\multicolumn{2}{l}{\textbf{By answer type}}                                                                                          \\ \hline
Entity              & \textit{\begin{tabular}[c]{@{}l@{}}Which country visited Japan in 2013?\end{tabular}} \\ \hline
Time             & \textit{\begin{tabular}[c]{@{}l@{}}When did China express intent to meet\\ with the Government of Pakistan?\end{tabular}}        \\ \bottomrule
\end{tabular}%
\caption{Representative examples from \ourq.}
\label{tab:examples}
\end{table}

\begin{figure*}[ht]
  \centering
  \includegraphics[width=\linewidth]{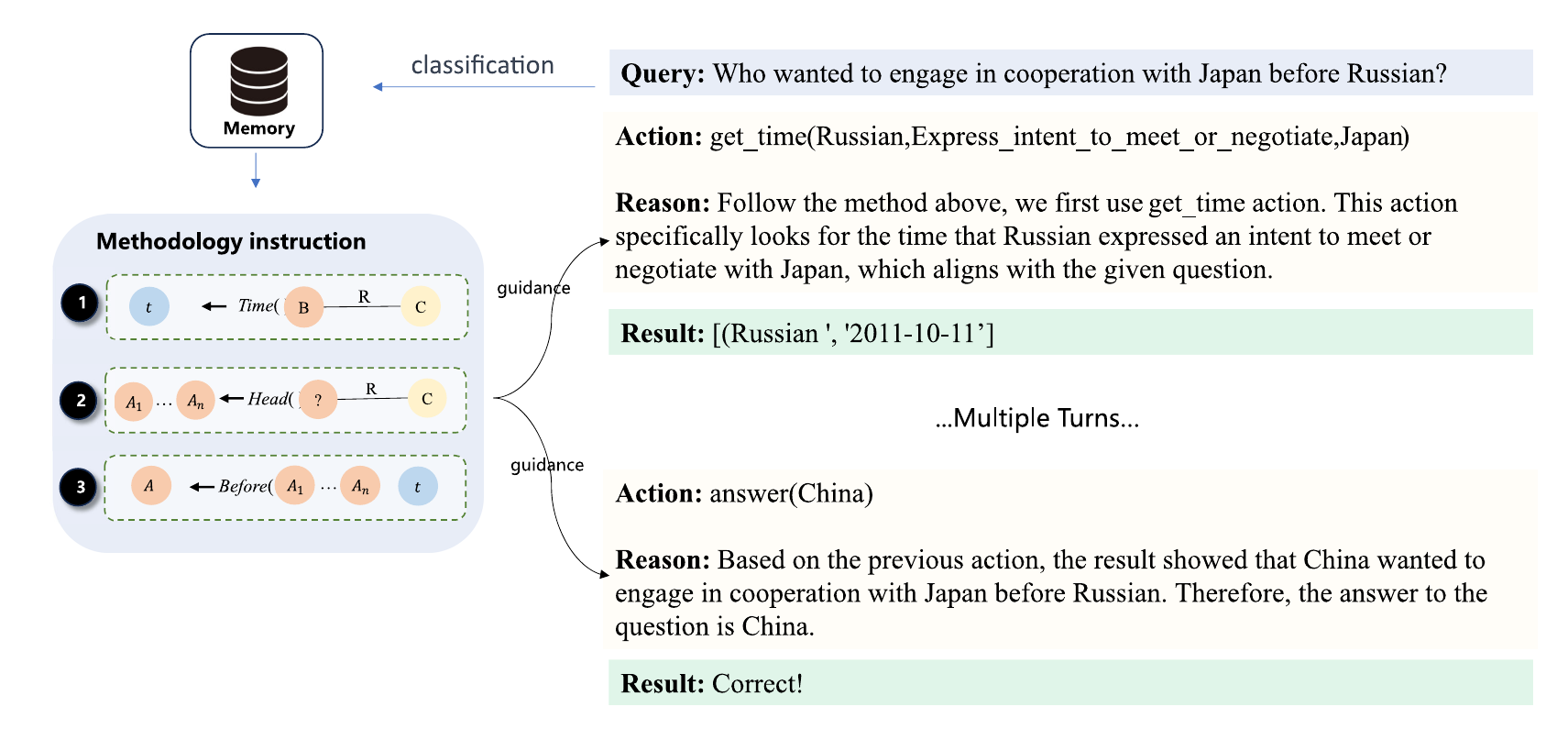}
  \caption{A demonstration sample of \ourmodel reasoning on \ourq.
  }
  \label{fig:ill}
 \vspace{-10pt}
\end{figure*}


\subsection{Action Templates}
\label{sec:actions}
Our approach is designed to be both generalizable and scalable. The templates in \ourmodel are not highly manually defined high-level functions, but rather finely-grained atomic operations (e.g., extracting time, entity extraction, etc.). These atomic operations can be flexibly combined to generalize a wide range of complex actions, demonstrating their versatility and extensibility.

Our action templates in \ourmodel strictly follow the definition of functions in~\autoref{tab:operators}. We employ several specialized functions to facilitate precise information retrieval. The \textit{getTime} function retrieves the timing of specific events, based on given entities and relation. For temporal positioning, \textit{getBefore}, \textit{getAfter}, and \textit{getBetween} identify entities or events relative to specified time frames. In terms of entity queries, \textit{getTailEntity} and \textit{getHeadEntity} ascertain linked entities based on existing relation, with an optional time constraint. For queries targeting specific time instances, \textit{getFirst} and \textit{getLast} pinpoint entities with the earliest and latest occurrences, respectively. Responses are then articulated using the \textit{answer} function, providing a streamlined method for answering queries within the TKG.

Our method exhibits low coupling with templates, making it adaptable to new data and domains. The extensibility of atomic templates is straightforward, allowing for easy incorporation of additional templates as needed. For instance, if we were to extend our approach to handle spatio-temporal data questions, adding a spatial atomic operation would be a straightforward task without the need for significant modifications.

\subsection{Details about the Instruction Format}
\label{sec:instruct}

\begin{table*}[t]
    \centering
    \small
    \resizebox{1.0\linewidth}{!}{
    \begin{tabular}{cl}
    \toprule
     \textbf{Action template}  & \textbf{Comments} \\
     \midrule
     \texttt{getTailEntity({head},{rel},{time})}  & Identify the tail/object entity based on the head/subject entity and relation\\
     \texttt{getHeadEntity({tail},{rel},{time})}  & Identify the head/suject entity based on the tail/object entity and relation\\
     \texttt{getTime({head},{rel},{tail})} & Retrieve the time of a specific event based on the head entity, relation and tail entity \\
    \texttt{getBetween({entities},{Time1},{Time2}}& Identify entities/events that occurred between two specific times\\
     \texttt{getBefore({entities},{time})} & Identify entities/events that occurred before a given time\\
     \texttt{getAfte({entities},{time})r}& Identify entities/events that occurred after a given time\\
     \texttt{getFirst({entities},{time})}& Pinpoint entities with the earliest occurrence\\
     \texttt{getLast({entities},{time})}& Pinpoint entities with the latest occurrence\\
     \texttt{answer(entities/time)} & To provide your answer, use the \texttt{answer} function\\
    \bottomrule    
    \end{tabular}}
    \caption{Action templates in \ourmodel.We employ these specialized functions to facilitate precise information retrieval.}
    \label{tab:operators}
\end{table*}

In Figure~\ref{fig:ill}, we illustrate an example of reasoning using the \ourmodel model. Table~\ref{tab:case} shows some exemplars of \ourmodel. During each step of the process, the LLM receives guidance from abstract methods and selects the optimal action from available paths, continuing until it deems an answer has been sufficiently formulated or the maximum reasoning length is reached. Figure~\ref{lst:instruction} presents the complete set of instructions used in our experiments, comprising components such as task definition, functional interpretations of potential actions, the current temporal question under consideration, historical reasoning steps, available candidate actions for the current round, feedback from the previous round's action, and requirements for output formatting.

\begin{figure}[ht]
  \centering
  \includegraphics[width=\linewidth]{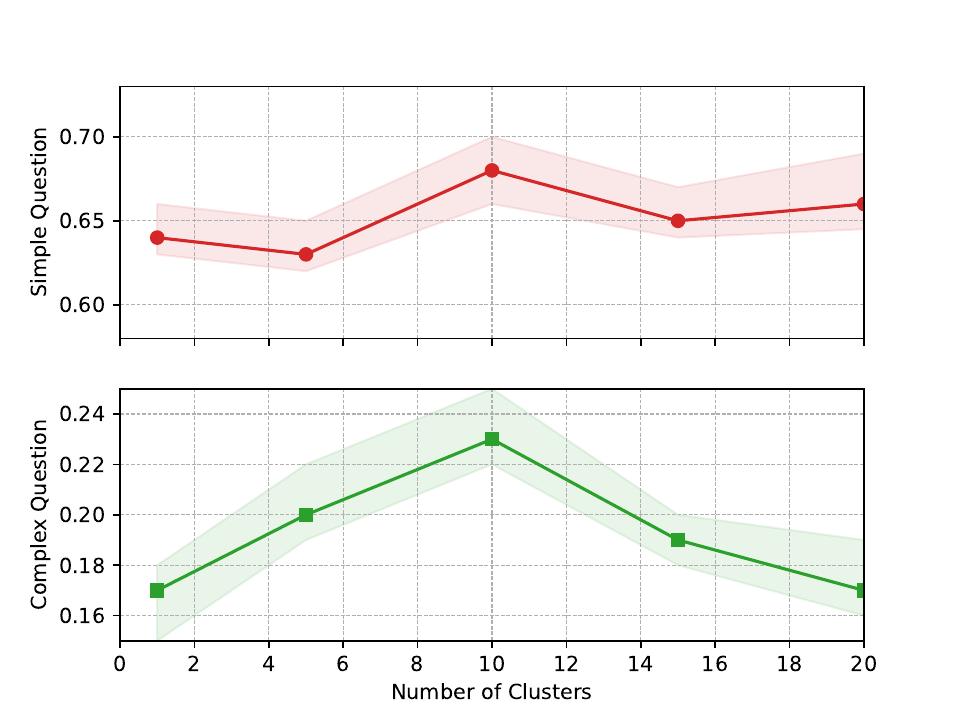}
  \caption{Accuracy v.s Number of Clusters of \ourmodel on \ourq.
  }
  \label{fig:cluster}
 \vspace{-10pt}
\end{figure}

\begin{figure}[ht]
  \centering
  \includegraphics[width=\linewidth]{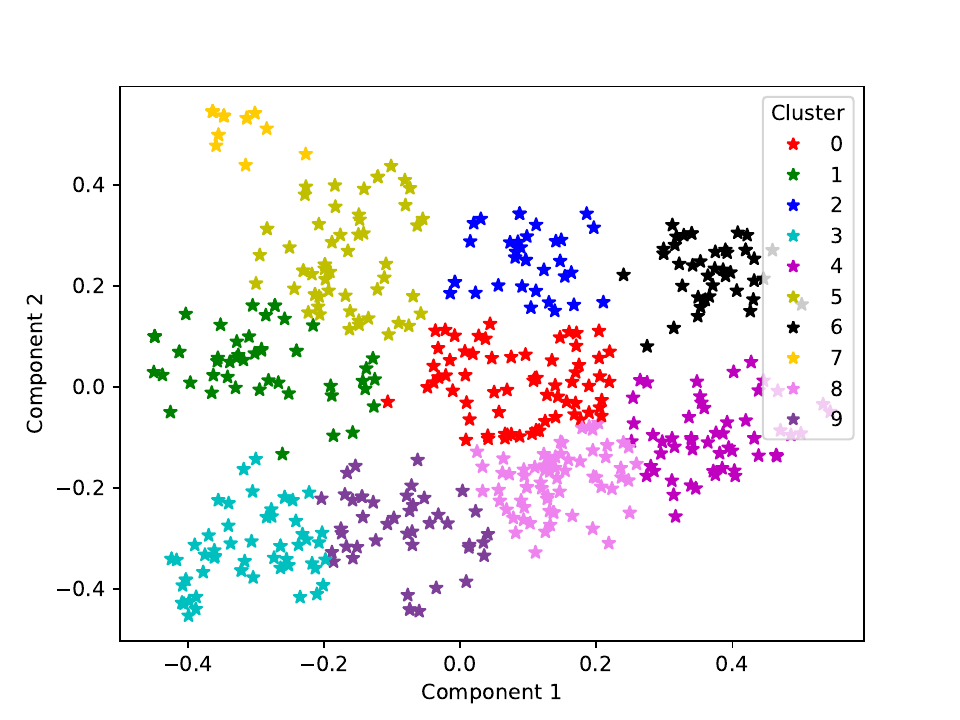}
  \caption{Clustering results for historical inference questions.  }
  \label{fig:cluster_pca}
  \vspace{-10pt}
\end{figure}

\subsection{Impact of Cluster Quantity} 
\label{sec:Cluster}
In Figure~\ref{fig:cluster_pca}, we show the reduced dimensional clustering diagram for the 10 categories of questions in the experiment. To verify the effect of different number of clusters on the results, we present the impact of the number of historical reasoning process clusters on the results. As shown in Figure \ref{fig:cluster}. We observe an initial increase followed by a decline in performance for both simple and complex problems. This pattern can be attributed to the fact that when the number of clusters is too low, the LLM is unable to distill concise and effective abstract methods from the noisy and abundant historical paths. Conversely, when the number of clusters is too high relative to a fixed number of historical samples, each category contains too few samples to provide the LLM with sufficient information to refine abstract methods. Thus, we observe a trend of improvement that eventually reverses as the number of clusters increases.

\subsection{Error Analysis}
For error analysis, we randomly sample 100 error instances from the test set and summarized the following three types of typical errors: (1) Retrieving irrelevant entities (in \ourq), meaning the model obtained wrong entities from the KG; Although our entity linking model can achieve a high prediction accuracy, wrong entities still exist in some questions. 
(2)Low-quality abstract methodological guidance. Within the dataset, there exist complex problems for which the historical reasoning processes consistently led to incorrect conclusions. This lack of sufficient correct reasoning histories hampers the LLM's ability to synthesize and refine effective abstract methodologies. Consequently, the low-quality abstract methods derived by the LLM prove inadequate in guiding subsequent reasoning processes, leading to a cascade of errors.
(3) Uncertainty outputs of LLMs. Despite the constraint that LLMs can only select from candidate actions or provide final answers, there are instances where they do not strictly adhere to the given instruction. This non-compliance leads to the failure of our predefined graph query methods, consequently impeding effective reasoning.

This demonstrates more efforts are needed to strengthen the model’s reasoning capability, particularly in the context of enhancing the reasoning capabilities of LLMs and diversifying their reasoning processes, it is crucial to provide a richer array of effective historical information for the generation of abstract methods. This approach is vital to prevent the LLMs from falling into a repetitive cycle of errors.

\subsection{Generalizability of Method on Other LLMs}
\label{sec:other_llm}
\begin{table}[ht]
\centering \small
\begin{tabular}{c|c|c}
\toprule
\multirow{1}{*}{\textbf{Model}} & 
\multicolumn{1}{c|}{\textbf{LLM only}} &  
\textbf{\ourmodel} \\ \cline{1-3}

\multicolumn{1}{c|}{\textbf{Llama-2 Chat 7B}} & 0.040 &0.105\\ 
\multicolumn{1}{c|}{\textbf{GPT-4}} & 0.125 &0.411\\ 
 \bottomrule
\end{tabular}%
\caption{Accuracy of \ourmodel with Other LLMs on \ourq}
\end{table}

To assess the effectiveness of the \ourmodel across various LLMs, we conducted experiments using the open-source model \texttt{Llama-2-7B-chat} and \texttt{GPT-4}. Our findings indicate that models with greater inherent capabilities yield better direct inference outcomes, presumably due to the acquisition of more extensive knowledge during training. Furthermore, the performance enhancements in \ourmodel models built upon these base models are more pronounced. Notably, the \ourmodel model based on \texttt{GPT-4} achieved a score of 0.411 on the \ourq benchmark, representing a relative improvement of 40.2\% over the current state-of-the-art models. This significant advancement underscores our model's robust generalization capabilities.

Our work focuses on enhancing LLM reasoning with abstract guidance, minimizing irrelevant noise by isolating them from direct knowledge sources. Unlike mainstream LLM agents designed for interactive environments, we employ a knowledge-agnostic module for environmental interactions, allowing the LLM to make decisions based on abstract guidance alone. 
Additionally, the study is open to incorporating interactive decision-making elements from recent LLM advancements, aiming to enhance the project's effectiveness and broaden its applicability in future research.


\begin{table*}[htbp]
\centering 
\begin{tabular}{ll}
\hline
\multicolumn{2}{l}{\textbf{Question 1}: In which month did the City Mayor of Philippines first praise Ona?} \\
\hline
\texttt{Action 0} & \texttt{getTime(City\_Mayor\_(Philippines),Praise\_or\_endorse,Ona)} \\
\texttt{Response 0} & \texttt{entities = [('City\_Mayor\_(Philippines)', '2014-10-07')]} \\
\texttt{Action 1} & \texttt{getFirst(\{entities\}) }\\
\texttt{Response 1} & \texttt{entities = [('City\_Mayor\_(Philippines)', '2014-10-07')]} \\
\texttt{Action 2} & \texttt{answer(2014-10)}\\
\texttt{Response 2} & \texttt{Correct!} \\ \hline

\multicolumn{2}{l}{\textbf{Question 2}: Before Mali's militant, which country was the last to criticise France?} \\
\hline
\texttt{Action 0} & \texttt{getTime(Militant\_(Mali),Criticize\_or\_denounce,France)} \\
\texttt{Response 0} & \texttt{entities = [('Militant\_(Mali)', '2013-01-21')]} \\
\texttt{Action 1} & \texttt{getFirst(\{entities\})}\\
\texttt{Response 1} & \texttt{entities = [("Government\_(Cote\_d'Ivoire)", ...]} \\
\texttt{Action 2} & \texttt{getLast(\{entities\})}\\
\texttt{Response 2} & \texttt{entities = [('Royal\_Administration\_(Spain)', '2015-12-24')]} \\
\texttt{Action 3} & \texttt{answer(Royal\_Administration\_(Spain))}\\
\texttt{Response 3} & \texttt{Correct!} \\
\hline

\multicolumn{2}{l}{\textbf{Question 3}: Who was the last to give a criticism to Iran before Pervez Musharraf did?} \\
\hline
\texttt{Action 0} & \texttt{getTime(Pervez\_Musharraf,Criticize\_or\_denounce,Iran)} \\
\texttt{Response 0} & \texttt{entities = [('Pervez\_Musharraf', '2006-01-17')]} \\
\texttt{Action 1} & \texttt{getHeadEntity(Iran,Criticize\_or\_denounce,no time)}\\
\texttt{Response 1} & \texttt{entities = [('Shirin\_Ebadi', '2005-01-05'), ...]} \\
\texttt{Action 2} & \texttt{getBefore(\{entities\},2006-01-17)}\\
\texttt{Response 2} & \texttt{entities = [('Shirin\_Ebadi', '2005-01-05'),...]} \\
\texttt{Action 3} & \texttt{getLast(\{entities\})}\\
\texttt{Response 3} & \texttt{entities = [('Angela\_Merkel', '2006-01-16')]} \\
\texttt{Action 4} & \texttt{answer(Angela\_Merkel)}\\
\texttt{Response 4} & \texttt{Correct!} \\
\hline

\multicolumn{2}{l}{\textbf{Question 4}: At what time did Xi Jinping first make optimistic remarks on Japan?} \\
\hline
\texttt{Action 0} & \texttt{getTime(Xi\_Jinping,Make\_optimistic\_comment,Japan)} \\
\texttt{Response 0} & \texttt{entities = [('Xi\_Jinping', '2008-04-18'), ...]} \\
\texttt{Action 1} & \texttt{getFirst(\{entities\})}\\
\texttt{Response 1} & \texttt{entities = [('Xi\_Jinping', '2008-04-18')]} \\
\texttt{Action 2} & \texttt{answer(2008-04-18)}\\
\texttt{Response 2} & \texttt{Correct!} \\
\hline

\multicolumn{2}{l}{\textbf{Question 5}: Who wanted to cooperate with Japan in November, 2005?} \\
\hline
\texttt{Action 0} & \texttt{getHeadEntity(Japan,Express\_intent\_to\_cooperate,2005-11)} \\
\texttt{Response 0} & \texttt{entities = [('Government\_Official\_(Russia)',  ...]} \\
\texttt{Action 1} & \texttt{answer(South\_Korea)}\\
\texttt{Response 1} & \texttt{Correct!} \\
\hline

\end{tabular}
\caption{ Exemplars of \ourmodel for \ourq}
\label{tab:case}
\end{table*}

\begin{figure*}[!ht]
\begin{lstlisting}[language=Python]
f'''Please use the tool provided below to interact with the knowledge graph. You will find a list of actions categorized into time-based queries, entity queries, and specific time queries. There may be more than one answer to the question, but you only need to answer one correct answer that satisfies the question.

To solve this question, you need to first identify the entities and relationships in the question, selecting the appropriate actions to retrieve the required information, and finally, providing the correct answer. 

Time-based Queries:
Retrieve the time of a specific event based on the head/subject entity, relation and tail/object entity by using the $get_time(HEAD, RELATION, TAIL)$ function, .
Identify entities/events that occurred before a given time by using the $get_before(ENTITY_LIST, SPECIFIED_TIME)$ function.
Identify entities/events that occurred after a given time by using the $get_after(ENTITY_LIST, SPECIFIED_TIME)$ function.
Identify entities/events that occurred between two specific times by using the $get_between(ENTITY_LIST, START_TIME, END_TIME)$ function.

Entity Queries:
Identify the tail/object entity based on the head/subject entity and relation by using the $get_tail_entity(CURRENT_HEAD, RELATION, OPTIONAL_TIME_CONSTRAINT)$ function.
Identify the head/suject entity based on the tail/object entity and relation by using the $get_head_entity(CURRENT_TAIL, RELATION, OPTIONAL_TIME_CONSTRAINT)$ function.

Specific Time Queries:
Pinpoint entities with the earliest occurrence by using the $get_first(ENTITY_LIST)$ function.
Identify entities with the latest occurrence by using the $get_last(ENTITY_LIST)$ function.
To provide your answer, use the $answer(YOUR_ANSWER)$ function.

Note: Always enclose the selected action in $ and provide a reason for your choice if necessary.

Examples for your reference: {examples}
(end of examples)

Current Challenge:

Question: {question}

Methodology: {methodology}
(end of methodology)

Previous Actions: {history}
(end of previous actions)

Available Actions: {actions}

Choose your next action from the available actions above, ensuring its completeness. If you have found the answer, remember to use the answer function.

Organize your output by strictly following the format below:

Action:
<Choose your next action from the available actions above. Note: Always enclose the selected action in $. Replace {your specified time} with a specified time in the format YYYY or YYYY-MM or YYYY-MM-DD>

Reason:
<Explain the reason for choosing this action.>'''

\end{lstlisting}
\caption{Prompt for the action selection.}
\label{lst:instruction}
\end{figure*}

\begin{figure*}[!ht]
\begin{lstlisting}[language=Python]
f'''Carefully analyze the following correct and incorrect examples. From these, extract and summarize the corresponding patterns and principles. Based on these examples, provide a comprehensive methodology that describes how to correctly tackle this type of problem, highlighting the key steps and common pitfalls to avoid.

Task Defination: <Task Defination>
(end of Task Defination) 

Here is an example output:
Example 1:
Overall methodology Instruction: 
This type of problem involves the sequential determination of events, e.g. Who {Relation R} {entity C} before {entity B}, to find the answer {entity A} we need to reason in three steps, firstly to determine the specific temporal anchors, i.e., the occurrence time t of {entity B, Relation, and entity C}, and then to find out which head entities have generated a Relation R connection with {entity C}. Then, we find out which head entities and {entity C} have been associated with Relation R, and finally filter out the answers that satisfy the time requirement before t. The specific steps are as follows. The steps are as follows

Step-by-step Guide:
1. Firstly, use get_time to find the time, $get_time(entity B, Relation R, entity C)$, to get the quaternion {entity B, Relation R, entity C, Time t};
2. use the get_head_entity method to get the head entity, $get_head_entity(entity C, Relation R, entity C)$, to be able to get a list of quaternions;
3. use the get_before method to filter the entities that satisfy the constraints, $get_before({entities},t)$, to be able to obtain a list of entities that satisfy the conditions
4. complete the reasoning process by answering the found answer $answer(entity A)$

(end of example output)

Here is the correct samples and incorrect samples for the current question type:
Correct samples: 
{correct_examples}

Incorrect samples:
{incorrect_examples}
(end of samples)

Now start writing. Please design a methodology that describes how to correctly tackle this type of problem. The goal is to provide a comprehensive guide that highlights the key steps and common pitfalls to avoid when approaching this type of problem.organize your output by strictly following the output format as below:

Overall Instruction: 
<Define this methodology in detail.  Provide a concise guide or inference. Note that the guidance you provide should be at a methodological level, for this type of question, not for a specific one. >

Step-by-step Guide:
<A step-by-step guide or procedure detailing how to approach and solve this kind of question. Note that the steps proposed should be specific and relevant to this type of question, tell which type of action should use in each step and the reason>'''

\end{lstlisting}
\caption{Prompt for the abstract methodology instruction generation.}
\label{lst:summary_instruction}
\end{figure*}